\documentclass[11pt,a4paper]{article}
\usepackage[hyperref]{acl2018}
\usepackage{times}
\usepackage{latexsym}
\usepackage{enumitem}
\usepackage{graphicx}
\usepackage{algorithm}
\usepackage{amsmath}
\usepackage{mathtools}
\usepackage{algorithmicx}

\usepackage{url}

\aclfinalcopy 

\title{Discourse Parsing in Videos: A Multi-modal Appraoch}

\author{Arjun R. Akula \\
  University of California, Los Angeles \\
  {\tt aakula@ucla.edu} \\\And
   Song-Chun Zhu \\
  University of California, Los Angeles \\
  {\tt sczhu@stat.ucla.edu} \\}

\date{}
\begin{document}

\maketitle
\begin{abstract}
Text-level discourse parsing aims to unmask how two segments (or sentences) in the text are related to each other. We propose the task of Visual Discourse Parsing, which requires understanding discourse relations among scenes in a video. Here we use the term scene to refer to a subset of video frames that can better summarize the video. In order to collect a dataset for learning discourse cues from videos, one needs to manually identify the scenes from a large pool of video frames and then annotate the discourse relations between them. This is clearly a time consuming, expensive and tedious task. In this work, we propose an approach to identify discourse cues from the videos without the need to explicitly identify and annotate the scenes. We also present a novel dataset containing 310 videos and the corresponding discourse cues to evaluate our approach. We believe that many of the multi-discipline Artificial Intelligence problems such as Visual Dialog and Visual Storytelling would greatly benefit from the use of visual discourse cues. Our code is publicly available at this github link: \url{https://github.com/arjunakula/Visual-Discourse-Parsing}
\end{abstract}

\section{Introduction}
Discourse structure aids in understanding a piece of text by linking it with other text units (such as surrounding clauses, sentences, etc.) from its context~\cite{carlson2003building,soricut2003sentence,lethanh2004generating}. A text span may be linked to another span through semantic relationships such as contrast relation, causal relation, etc.~\cite{marcu2002unsupervised, duverle2009novel}. Text-level discourse parsing algorithms aim to unmask such relationships in text, which is central to many downstream natural language processing (NLP) applications such as information retrieval, text summarization, sentiment analysis~\cite{wang2015refined} and question answering~\cite{chai2004discourse,akula2013novel,akula2015novel}. 

\begin{figure*}
\centering
  \includegraphics[width=1.7\columnwidth, height=0.9\columnwidth]{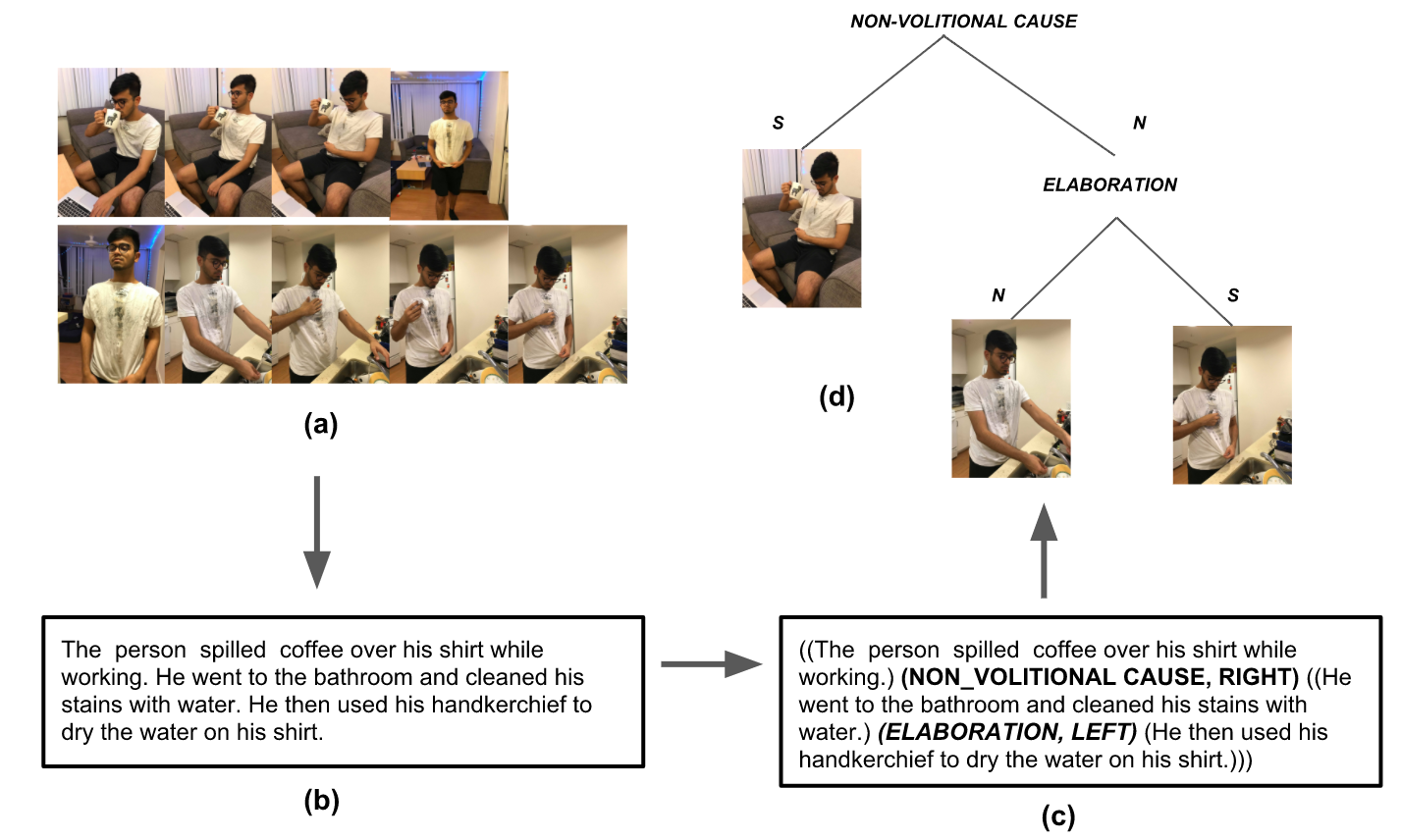}
  \caption{(a) Video with 9 frames; (b) Textual description of video; (c) RST discourse structure (represented as sequence of words) of description in (b). The notations LEFT and RIGHT represent the direction (nuclearity) of the rhetorical relations; (d) RST discourse structure of the video using 3 frames, i.e. scenes. The symbols N and S indicate the nucleus and satellite of each rhetorical relation.}~\label{fig:figure1}
\end{figure*}

Recently, there has been a lot of focus on multi-discipline Artificial Intelligence (AI) research problems such as visual storytelling~\cite{huang2016visual} and visual dialog~\cite{das2016visual}. Solving these problems requires multi-modal knowledge that combines computer vision (CV), NLP, and knowledge representation \& reasoning (KR), making the need for commonsense knowledge and complex reasoning more essential. In an effort to fill this need, we introduce the task of Visual Discourse Parsing.

\noindent \textbf{Task Definition.} The concrete task in Visual Discourse Parsing is the following - given a video, understand discourse relationships among its scenes. Specifically, given a video, the task is to identify a scene's relation with the context. Here we use the term scene to refer to a subset of video frames that can better summarize the video. We use Rhetorical Structure Theory (RST)~\cite{mann1988rhetorical} to capture discourse relations among the scenes~\cite{DBLP:journals/corr/abs-1903-02252,carlson2003building,soricut2003sentence,lethanh2004generating}.

Consider for example, nine frames of a video shown in Figure~\ref{fig:figure1}. We can represent the discourse structure of this video using only 3 out of 9 frames, i.e. there are only 3 scenes for this video. The discourse structure in the Figure~\ref{fig:figure1} interprets the video as follows: the event ``person going to the bathroom and cleaning his stains" is \textit{caused by} the event ``the person spilling coffee over his shirt"; the event ``the person used his handkerchief to dry the water on his shirt" is simply an \textit{elaboration} of the event ``person going to the bathroom and cleaning his stains"~\cite{DBLP:journals/corr/abs-1903-02252,akula20words,akula2019visual,akula2021crossvqa,akula2021measuring,akula2021robust,akula2020cocox,r2019natural,pulijala2013web,gupta2012novel}. 

It is a time consuming, expensive and tedious task to collect a dataset for learning discourse cues from videos. This is because one needs to manually identify the scenes from a large pool of video frames and annotate the discourse relations between them. To this end, we propose an approach to identify discourse cues from the videos without the need to explicitly identify and annotate the scenes. We also present a novel dataset containing 310 videos and the corresponding discourse cues to evaluate our approach~\cite{akula2013novel,akula2018analyzing,akula2021mind,gupta2016desire,akula2019explainable,akula2021gaining,akula2019x,akula2020words}.

\section{Approach}
 Algorithm~\ref{alg:the_alg} presents our approach for learning a model to identify discourse structure from the videos. Firstly, we generate natural language text descriptions from the videos automatically using video captioning methods such as~\cite{yu2016video},~\cite{venugopalan2016improving} and~\cite{pasunuru2017multi}. Secondly, we obtain discourse structures of the above text descriptions using text-level discourse parsers such as~\cite{duverle2009novel} and~\cite{ji2014representation}. We represent the discourse structure as a sequence of words (see Figure~\ref{fig:figure1}). 
 
 Next we use an end-to-end trainable architecture for learning to predict the text-level discourse structures (i.e. sequence of words) from the videos (i.e. sequence of video frames). This gives us a model to map videos to their corresponding text-level discourse structures. Finally we use saliency methods~\cite{ramanishka2017top} to replace the textual descriptions in the discourse structure, i.e. elementary discourse units (EDUs), with the scenes. For example, in Figure~\ref{fig:figure1}, \textbf{(b)} is the text description of video shown in \textbf{(a)}. Text-level discourse structure of \textbf{(b)} is shown in \textbf{(c)} as a sequence of words. We then map \textbf{(c)} to \textbf{(d)} using saliency methods. 
 
\begin{algorithm}
\caption{Our approach for Visual Discourse Parsing.}
\begin{algorithmic}[1]
\State Generate natural language descriptions of videos.
\State Find discourse structure of text descriptions. 
\State Train a sequence to sequence model to learn the mapping from video frames to the sequence of words in the discourse structure.
\State Use saliency methods to map EDUs in text-level discourse structure to scenes.
\end{algorithmic}
\label{alg:the_alg}
\end{algorithm}

The quality of natural language descriptions and the their text-level discourse structures is crucial for learning a robust model. The state-of-the-art video captioning and text-level discourse parsing approaches, as we found in our experiments, may generate a lot of noise in their outputs. While developing our corpus, we manually performed these two steps. These manual annotations are still much easier to perform compared to the tedious task of directly annotating discourse relations between video frames.

In the step 3 of our algorithm, we use the standard machine translation encoder-decoder RNN model~\cite{sutskever2014sequence}. As RNN suffers from decaying of gradient and blowing-up of gradient problem, we use LSTM units, which are good at memorizing long-range dependencies due to forget-style gates~\cite{hochreiter1997long}. The sequence of video frames are passed to the encoder. The last hidden state of the encoder is then passed to the decoder. The decoder generates the discourse structure as a sequence of words. Let the input sequence of video frames be $\mathbf{x} = \left(x\_1, ... ,x\_p\right)$ and the output sequence of words as $\mathbf{y} = \left(y\_1,...,y\_n\right)$. The distribution of the output sequence w.r.t. the input sequence is:

\begin{equation}
p(y_1,...,y_n \mid x_1,...,x_p) = \prod_{t=1}^{n} p(y_t\mid h_t^d)
\end{equation}
where $h_t^d$ is the hidden state at the $t^{th}$ time step of the decoding LSTM.

\noindent \textbf{Soft Attention:} We further improve our encoder-decoder model using an attention based sequence-to-sequence model~\cite{bahdanau2014neural}. The attention weights act as an alignment mechanism by re-weighting the encoder hidden states that are more relevant for decoder time step.

\begin{table*}
  \centering
  \begin{tabular}{|l| r| r| r| r| r| r| r|}
    \hline
    {\small\textit{RNN Type}}
    & {\small \textit{\#Hidden Units}}
      & {\small \textit{Bidirectional}}
    & {\small \textit{\#Layers}} 
    & {\small \textit{Relations}} 
    & {\small \textit{Edges}}
    & {\small \textit{Relations+Edges}}
    & {\small \textit{Bleu4}} \\

    \hline
    LSTM & 256 & NO & 1 & 0.3 & 0.51 & 0.21 & 0.22\\
LSTM & 512 & NO & 1 & 0.52 & 0.62 & 0.42 & 0.41\\
LSTM & 1024 & YES & 1 & 0.49 & 0.51 & 0.42 & 0.33\\
LSTM & 1024 & NO & 1 & 0.35 & 0.51 & 0.21 & 0.34\\
LSTM & 512 & NO & 2 & 0.35 & 0.51 & 0.21 & 0.38\\
LSTM & 512 & NO & 3 & 0.56 & 0.62 & 0.42 & 0.39\\
LSTM & 512 & NO & 4 & 0.56 & 0.62 & 0.42 & 0.39\\
GRU & 512 & NO & 1 & 0.3 & 0.51 & 0.21 & 0.33\\
\hline
  \end{tabular}
  \caption{Evaluation using sequence-to-sequence model without Attention.}~\label{tab:table1}
\end{table*}

\begin{table*}
  \centering
  \begin{tabular}{| l | r | r| r| r| r| r| r| r|}
    \hline
    {\small\textit{RNN Type}}
    & {\small \textit{\#Hidden Units}}
      & {\small \textit{Bidirectional}}
    & {\small \textit{\#Layers}}
    & {\small \textit{\#Attention Type}} 
    & {\small \textit{Relations}} 
    & {\small \textit{Edges}}
    & {\small \textit{Relations+Edges}}
    & {\small \textit{Bleu4}} \\
   \hline
    LSTM & 512 & NO & 1 & general & 0.63 & 0.69 & 0.53 & 0.59\\
LSTM & 512 & NO & 1 & dot & 0.52 & 0.65 & 0.45 & 0.52\\
LSTM & 512 & NO & 1 & concat & 0.52 & 0.65 & 0.45 & 0.51\\
LSTM & 512 & NO & 2 & general & 0.52 & 0.65 & 0.45 & 0.47\\
LSTM & 512 & NO & 3 & general & 0.5 & 0.65 & 0.39 & 0.41\\
\hline
  \end{tabular}
  \caption{Evaluation using sequence-to-sequence model using Attention.}~\label{tab:table2}
\end{table*}
\section{Experiments}
We developed a new dataset containing 310 videos. These videos are shot at various settings such as playing sports (Table Tennis, Frisbee, Tennis, Rugby), bus stop, dining hall, elevator, classroom, library, garden, study room, etc. On average, the length of each video is about 19 seconds. We first manually generated descriptions of each video and then annotated the discourse structure of these descriptions - with the help of 5 graduate students. Each video is annotated by at least 2 students. We solved the disagreements found in the annotations together. As the training data is not large enough, we chose short videos and described each video using only three sentences, i.e. discourse structure (RST tree) of each video contains only two relations and two edges. This reduces the total number of parameters that need to be learned from the end-to-end training (in step 3 of Algorithm~\ref{alg:the_alg}).

We evaluate our approach by using the following three metrics:
\begin{enumerate}[label=(\alph*)]
\item \textbf{BLEU score} We used the BLEU score~\cite{papineni2002bleu} to evaluate the translation quality of the discourse structure generated from the videos. We computed our BLEU score on the tokenized predictions and ground truth. 
\item \textbf{Relations Accuracy} Each video, in our dataset, contains two discourse relations. The Relations Accuracy metric is defined as the total number of relations correctly predicted by the model. 
\item \textbf{Edges Accuracy} Each video, in our dataset, contains two edges. The Edges Accuracy metric is defined as the total number of edges (i.e. RST node nuclearity directions) correctly predicted by the model.
\item \textbf{Relations+Edges Accuracy} Here, we compute the correctness of the complete discourse structure, i.e. the predicted discourse structure will be considered correct only if all the relations and the edges are correctly predicted by the model.  
\end{enumerate}

\noindent
We use 210 videos for training, 30 videos for validation and the rest 70 videos for testing. We fix our sampling rate to 5fps to bring uniformity in the temporal representation of actions across all videos. These sampled frames are then converted into features using VGGNet~\cite{simonyan2014very}. We initialize the decoder embedding with Google pre-trained word2vec word embeddings~\cite{mikolov2013distributed}. We tune all hyperparameters using our validation data: learning rate, weight initializations, hidden states. We use a 1024-dimension RNN hidden state size. We use Adam optimizer~\cite{kingma2014adam} and apply a dropout of 0.5.

\noindent
Table~\ref{tab:table1} presents our results, reporting several configurations of the encoder-decoder RNN model, using the four evaluation metrics. The metrics Relations, Edges and Relations+Edges accuracies are scaled between 0 and 1. It may be noted that RNN configuration using LSTM unit with 512 hidden units and 3,4 encoder layers outperformed other RNN configurations. Table~\ref{tab:table2} reports the performance of our attention based sequence to sequence model. Attention based models gave better accuracies than the simple sequence to sequence models. In particular, the LSTM unit with 512 hidden units and single encoding layer outperformed other configurations. These results are encouraging considering the small size of our training dataset.

\section{Related Work}
\textbf{Text-level Discourse Parsing}\\
Several works~\cite{carlson2003building,lethanh2004generating, marcu2002unsupervised, duverle2009novel} have been proposed in the past to parse text documents into RST style discourse representation. ~\cite{carlson2003building} proposed a data driven approach using RST Discourse Treebank. ~\cite{reitter2003simple} proposed a chart-parsing-style techniques to derive discourse trees from documents~\cite{akula2022effective,agarwal2018structured,akula2019natural,akula2015novel,palakurthi2015classification,agarwal2017automatic,dasgupta2014towards}. ~\cite{lethanh2004generating} proposed a three-stage approach to find discourse parse. It segments the elementary discourse units (EDUs) into trees for each successive level of document: first at the sentence level, then paragraph and finally at the document level. However, using a three-stage approach exponentially increased their search space, making it computationally intractable to find the optimal discourse tree. ~\cite{duverle2009novel} proposed end-to-end parsing algorithms. However, in none of these works, the problem of extracting discourse information from videos has been addressed.\\
\textbf{Video Captioning}\\
There are few works on video captioning which come closer to our line of research. Video captioning is the task of describing the content of a video. ~\cite{guadarrama2013youtube2text,thomason2014integrating,yu2016video} proposed a multi-stage approach to identify dependency roles such as subject and object in order to generate a description of the video. More recently,~\cite{venugopalan2016improving} proposed a sequence-to-sequence model with encoder and decoder RNNs. ~\cite{pasunuru2017multi} proposed a multi-task learning approach to generate captions from the video. However, all of these works mainly focus on generating text from the video but not on understanding relationships between the scenes (or frames).

\section{Conclusions}
We introduced a new AI task - Visual Discourse Parsing, where the AI agent needs to understand discourse relations among scenes in a video. We presented an end-to-end learning approach to identify the discourse structure of videos. Central to our approach is the use of text descriptions of videos to identify discourse relations. In the future, we plan to extend this dataset to include longer videos that need more than three sentences to describe. We also intend to experiment with multi-task learning approaches. Our results indicate that there is significant scope for improvement.


\begin{thebibliography}{}
\expandafter\ifx\csname natexlab\endcsname\relax\def\natexlab#1{#1}\fi

\bibitem[{Agarwal et~al.(2017)Agarwal, Aggarwal, Akula, Dasgupta, and
  Sridhara}]{agarwal2017automatic}
Shivali Agarwal, Vishalaksh Aggarwal, Arjun~R Akula, Gargi~Banerjee Dasgupta,
  and Giriprasad Sridhara. 2017.
\newblock Automatic problem extraction and analysis from unstructured text in
  it tickets.
\newblock {\em IBM Journal of Research and Development\/} 61(1):4--41.

\bibitem[{Agarwal et~al.(2018)Agarwal, Akula, Dasgupta, Nadgowda, and
  Nayak}]{agarwal2018structured}
Shivali Agarwal, Arjun~R Akula, Gaargi~B Dasgupta, Shripad~J Nadgowda, and
  Tapan~K Nayak. 2018.
\newblock Structured representation and classification of noisy and
  unstructured tickets in service delivery.
\newblock US Patent 10,095,779.

\bibitem[{Akula et~al.(2021{\natexlab{a}})Akula, Gella, Wang, Zhu, and
  Reddy}]{akula2021mind}
Arjun Akula, Spandana Gella, Keze Wang, Song-chun Zhu, and Siva Reddy.
  2021{\natexlab{a}}.
\newblock Mind the context: The impact of contextualization in neural module
  networks for grounding visual referring expressions.
\newblock In {\em Proceedings of the 2021 Conference on Empirical Methods in
  Natural Language Processing\/}. pages 6398--6416.

\bibitem[{Akula et~al.(2021{\natexlab{b}})Akula, Jampani, Changpinyo, and
  Zhu}]{akula2021robust}
Arjun Akula, Varun Jampani, Soravit Changpinyo, and Song-Chun Zhu.
  2021{\natexlab{b}}.
\newblock Robust visual reasoning via language guided neural module networks.
\newblock {\em Advances in Neural Information Processing Systems\/} 34.

\bibitem[{Akula et~al.(2013)Akula, Sangal, and Mamidi}]{akula2013novel}
Arjun Akula, Rajeev Sangal, and Radhika Mamidi. 2013.
\newblock A novel approach towards incorporating context processing
  capabilities in nlidb system.
\newblock In {\em Proceedings of the sixth international joint conference on
  natural language processing\/}. pages 1216--1222.

\bibitem[{Akula and Zhu(2022)}]{akula2022effective}
Arjun Akula and Song-Chun Zhu. 2022.
\newblock Effective representation to capture collaboration behaviors between
  explainer and user.
\newblock {\em arXiv preprint arXiv:2201.03147\/} .

\bibitem[{Akula(2015)}]{akula2015novel}
Arjun~R Akula. 2015.
\newblock A novel approach towards building a generic, portable and contextual
  nlidb system.
\newblock {\em International Institute of Information Technology Hyderabad\/} .

\bibitem[{Akula et~al.(2021{\natexlab{c}})Akula, Changpinyo, Gong, Sharma, Zhu,
  and Soricut}]{akula2021crossvqa}
Arjun~R Akula, Beer Changpinyo, Boqing Gong, Piyush Sharma, Song-Chun Zhu, and
  Radu Soricut. 2021{\natexlab{c}}.
\newblock Crossvqa: Scalably generating benchmarks for systematically testing
  vqa generalization .

\bibitem[{Akula et~al.(2018)Akula, Dasgupta, and Nayak}]{akula2018analyzing}
Arjun~R Akula, Gaargi~B Dasgupta, and Tapan~K Nayak. 2018.
\newblock Analyzing tickets using discourse cues in communication logs.
\newblock US Patent 10,067,983.

\bibitem[{Akula et~al.(2021{\natexlab{d}})Akula, Dasgupta, Ekambaram, and
  Narayanam}]{akula2021measuring}
Arjun~R Akula, Gargi~B Dasgupta, Vijay Ekambaram, and Ramasuri Narayanam.
  2021{\natexlab{d}}.
\newblock Measuring effective utilization of a service practitioner for ticket
  resolution via a wearable device.
\newblock US Patent 10,929,264.

\bibitem[{Akula et~al.(2020{\natexlab{a}})Akula, Gella, Al-Onaizan, Zhu, and
  Reddy}]{akula20words}
Arjun~R. Akula, Spandana Gella, Yaser Al-Onaizan, Song-Chun Zhu, and Siva
  Reddy. 2020{\natexlab{a}}.
\newblock Words aren't enough, their order matters: On the robustness of
  grounding visual referring expressions.
\newblock In {\em ACL\/}.

\bibitem[{Akula et~al.(2020{\natexlab{b}})Akula, Gella, Al-Onaizan, Zhu, and
  Reddy}]{akula2020words}
Arjun~R Akula, Spandana Gella, Yaser Al-Onaizan, Song-Chun Zhu, and Siva Reddy.
  2020{\natexlab{b}}.
\newblock Words aren't enough, their order matters: On the robustness of
  grounding visual referring expressions.
\newblock {\em arXiv preprint arXiv:2005.01655\/} .

\bibitem[{Akula et~al.(2019{\natexlab{a}})Akula, Liu, Saba-Sadiya, Lu,
  Todorovic, Chai, and Zhu}]{akula2019x}
Arjun~R Akula, Changsong Liu, Sari Saba-Sadiya, Hongjing Lu, Sinisa Todorovic,
  Joyce~Y Chai, and Song-Chun Zhu. 2019{\natexlab{a}}.
\newblock X-tom: Explaining with theory-of-mind for gaining justified human
  trust.
\newblock {\em arXiv preprint arXiv:1909.06907\/} .

\bibitem[{Akula et~al.(2019{\natexlab{b}})Akula, Liu, Todorovic, Chai, and
  Zhu}]{akula2019explainable}
Arjun~R Akula, Changsong Liu, Sinisa Todorovic, Joyce~Y Chai, and Song-Chun
  Zhu. 2019{\natexlab{b}}.
\newblock Explainable ai as collaborative task solving.
\newblock In {\em CVPR Workshops\/}. pages 91--94.

\bibitem[{Akula et~al.(2019{\natexlab{c}})Akula, Todorovic, Chai, and
  Zhu}]{akula2019natural}
Arjun~R Akula, Sinisa Todorovic, Joyce~Y Chai, and Song-Chun Zhu.
  2019{\natexlab{c}}.
\newblock Natural language interaction with explainable ai models.
\newblock In {\em CVPR Workshops\/}. pages 87--90.

\bibitem[{Akula et~al.(2020{\natexlab{c}})Akula, Wang, and
  Zhu}]{akula2020cocox}
Arjun~R. Akula, Shuai Wang, and Song{-}Chun Zhu. 2020{\natexlab{c}}.
\newblock \href{https://aaai.org/ojs/index.php/AAAI/article/view/5643}{Cocox:
  Generating conceptual and counterfactual explanations via fault-lines}.
\newblock In {\em The Thirty-Fourth {AAAI} Conference on Artificial
  Intelligence, {AAAI} 2020, The Thirty-Second Innovative Applications of
  Artificial Intelligence Conference, {IAAI} 2020, The Tenth {AAAI} Symposium
  on Educational Advances in Artificial Intelligence, {EAAI} 2020, New York,
  NY, USA, February 7-12, 2020\/}. {AAAI} Press, pages 2594--2601.
\newblock
  \href{https://aaai.org/ojs/index.php/AAAI/article/view/5643}{https://aaai.org/ojs/index.php/AAAI/article/view/5643}.

\bibitem[{Akula and
  Zhu(2019{\natexlab{a}})}]{DBLP:journals/corr/abs-1903-02252}
Arjun~R. Akula and Song{-}Chun Zhu. 2019{\natexlab{a}}.
\newblock \href{http://arxiv.org/abs/1903.02252}{Visual discourse parsing}.
\newblock {\em CVPR 2019 Workshop on Language and Vision, arXiv:1903.02252\/}
  \href{http://arxiv.org/abs/1903.02252}{http://arxiv.org/abs/1903.02252}.

\bibitem[{Akula and Zhu(2019{\natexlab{b}})}]{akula2019visual}
Arjun~R Akula and Song-Chun Zhu. 2019{\natexlab{b}}.
\newblock \href{https://arxiv.org/abs/1903.02252}{Visual discourse parsing}.
\newblock {\em ArXiv preprint\/} abs/1903.02252.
\newblock
  \href{https://arxiv.org/abs/1903.02252}{https://arxiv.org/abs/1903.02252}.

\bibitem[{Akula(2021)}]{akula2021gaining}
Arjun~Reddy Akula. 2021.
\newblock {\em Gaining Justified Human Trust by Improving Explainability in
  Vision and Language Reasoning Models\/}.
\newblock Ph.D. thesis, UCLA.

\bibitem[{Bahdanau et~al.(2014)Bahdanau, Cho, and Bengio}]{bahdanau2014neural}
Dzmitry Bahdanau, Kyunghyun Cho, and Yoshua Bengio. 2014.
\newblock Neural machine translation by jointly learning to align and
  translate.
\newblock {\em arXiv preprint arXiv:1409.0473\/} .

\bibitem[{Carlson et~al.(2003)Carlson, Marcu, and
  Okurowski}]{carlson2003building}
Lynn Carlson, Daniel Marcu, and Mary~Ellen Okurowski. 2003.
\newblock Building a discourse-tagged corpus in the framework of rhetorical
  structure theory.
\newblock In {\em Current and new directions in discourse and dialogue\/},
  Springer, pages 85--112.

\bibitem[{Chai and Jin(2004)}]{chai2004discourse}
Joyce~Y Chai and Rong Jin. 2004.
\newblock Discourse structure for context question answering.
\newblock In {\em Proceedings of the Workshop on Pragmatics of Question
  Answering at HLT-NAACL 2004\/}.

\bibitem[{Das et~al.(2016)Das, Kottur, Gupta, Singh, Yadav, Moura, Parikh, and
  Batra}]{das2016visual}
Abhishek Das, Satwik Kottur, Khushi Gupta, Avi Singh, Deshraj Yadav,
  Jos{\'e}~MF Moura, Devi Parikh, and Dhruv Batra. 2016.
\newblock Visual dialog.
\newblock {\em arXiv preprint arXiv:1611.08669\/} .

\bibitem[{Dasgupta et~al.(2014)Dasgupta, Nayak, Akula, Agarwal, and
  Nadgowda}]{dasgupta2014towards}
Gargi~B Dasgupta, Tapan~K Nayak, Arjun~R Akula, Shivali Agarwal, and Shripad~J
  Nadgowda. 2014.
\newblock Towards auto-remediation in services delivery: Context-based
  classification of noisy and unstructured tickets.
\newblock In {\em International Conference on Service-Oriented Computing\/}.
  Springer, pages 478--485.

\bibitem[{Duverle and Prendinger(2009)}]{duverle2009novel}
David~A Duverle and Helmut Prendinger. 2009.
\newblock A novel discourse parser based on support vector machine
  classification.
\newblock In {\em Proceedings of the Joint Conference of the 47th Annual
  Meeting of the ACL and the 4th International Joint Conference on Natural
  Language Processing of the AFNLP: Volume 2-Volume 2\/}. Association for
  Computational Linguistics, pages 665--673.

\bibitem[{Guadarrama et~al.(2013)Guadarrama, Krishnamoorthy, Malkarnenkar,
  Venugopalan, Mooney, Darrell, and Saenko}]{guadarrama2013youtube2text}
Sergio Guadarrama, Niveda Krishnamoorthy, Girish Malkarnenkar, Subhashini
  Venugopalan, Raymond Mooney, Trevor Darrell, and Kate Saenko. 2013.
\newblock Youtube2text: Recognizing and describing arbitrary activities using
  semantic hierarchies and zero-shot recognition.
\newblock In {\em Proceedings of the IEEE international conference on computer
  vision\/}. pages 2712--2719.

\bibitem[{Gupta et~al.(2012)Gupta, Akula, Malladi, Kukkadapu, Ainavolu, and
  Sangal}]{gupta2012novel}
Abhijeet Gupta, Arjun Akula, Deepak Malladi, Puneeth Kukkadapu, Vinay Ainavolu,
  and Rajeev Sangal. 2012.
\newblock A novel approach towards building a portable nlidb system using the
  computational paninian grammar framework.
\newblock In {\em 2012 International Conference on Asian Language
  Processing\/}. IEEE, pages 93--96.

\bibitem[{Gupta et~al.(2016)Gupta, Akula, Dasgupta, Aggarwal, and
  Mohapatra}]{gupta2016desire}
Abhirut Gupta, Arjun Akula, Gargi Dasgupta, Pooja Aggarwal, and Prateeti
  Mohapatra. 2016.
\newblock Desire: Deep semantic understanding and retrieval for technical
  support services.
\newblock In {\em International Conference on Service-Oriented Computing\/}.
  Springer, pages 207--210.

\bibitem[{Hochreiter and Schmidhuber(1997)}]{hochreiter1997long}
Sepp Hochreiter and J{\"u}rgen Schmidhuber. 1997.
\newblock Long short-term memory.
\newblock {\em Neural computation\/} 9(8):1735--1780.

\bibitem[{Huang et~al.(2016)Huang, Ferraro, Mostafazadeh, Misra, Agrawal,
  Devlin, Girshick, He, Kohli, Batra et~al.}]{huang2016visual}
Ting-Hao~Kenneth Huang, Francis Ferraro, Nasrin Mostafazadeh, Ishan Misra,
  Aishwarya Agrawal, Jacob Devlin, Ross Girshick, Xiaodong He, Pushmeet Kohli,
  Dhruv Batra, et~al. 2016.
\newblock Visual storytelling.
\newblock In {\em Proceedings of the 2016 Conference of the North American
  Chapter of the Association for Computational Linguistics: Human Language
  Technologies\/}. pages 1233--1239.

\bibitem[{Ji and Eisenstein(2014)}]{ji2014representation}
Yangfeng Ji and Jacob Eisenstein. 2014.
\newblock Representation learning for text-level discourse parsing.
\newblock In {\em Proceedings of the 52nd Annual Meeting of the Association for
  Computational Linguistics (Volume 1: Long Papers)\/}. volume~1, pages 13--24.

\bibitem[{Kingma and Ba(2015)}]{kingma2014adam}
Diederik~P Kingma and Jimmy Ba. 2015.
\newblock Adam: A method for stochastic optimization.
\newblock {\em International Conference on Learning Representations (ICLR)\/} .

\bibitem[{LeThanh et~al.(2004)LeThanh, Abeysinghe, and
  Huyck}]{lethanh2004generating}
Huong LeThanh, Geetha Abeysinghe, and Christian Huyck. 2004.
\newblock Generating discourse structures for written texts.
\newblock In {\em Proceedings of the 20th international conference on
  Computational Linguistics\/}. Association for Computational Linguistics, page
  329.

\bibitem[{Mann and Thompson(1988)}]{mann1988rhetorical}
William~C Mann and Sandra~A Thompson. 1988.
\newblock Rhetorical structure theory: Toward a functional theory of text
  organization.
\newblock {\em Text-Interdisciplinary Journal for the Study of Discourse\/}
  8(3):243--281.

\bibitem[{Marcu and Echihabi(2002)}]{marcu2002unsupervised}
Daniel Marcu and Abdessamad Echihabi. 2002.
\newblock An unsupervised approach to recognizing discourse relations.
\newblock In {\em Proceedings of the 40th Annual Meeting on Association for
  Computational Linguistics\/}. Association for Computational Linguistics,
  pages 368--375.

\bibitem[{Mikolov et~al.(2013)Mikolov, Sutskever, Chen, Corrado, and
  Dean}]{mikolov2013distributed}
Tomas Mikolov, Ilya Sutskever, Kai Chen, Greg~S Corrado, and Jeff Dean. 2013.
\newblock Distributed representations of words and phrases and their
  compositionality.
\newblock In {\em Advances in neural information processing systems\/}. pages
  3111--3119.

\bibitem[{Palakurthi et~al.(2015)Palakurthi, Ruthu, Akula, and
  Mamidi}]{palakurthi2015classification}
Ashish Palakurthi, SM~Ruthu, Arjun Akula, and Radhika Mamidi. 2015.
\newblock Classification of attributes in a natural language query into
  different sql clauses.
\newblock In {\em Proceedings of the International Conference Recent Advances
  in Natural Language Processing\/}. pages 497--506.

\bibitem[{Papineni et~al.(2002)Papineni, Roukos, Ward, and
  Zhu}]{papineni2002bleu}
Kishore Papineni, Salim Roukos, Todd Ward, and Wei-Jing Zhu. 2002.
\newblock Bleu: a method for automatic evaluation of machine translation.
\newblock In {\em Proceedings of the 40th annual meeting on association for
  computational linguistics\/}. Association for Computational Linguistics,
  pages 311--318.

\bibitem[{Pasunuru and Bansal(2017)}]{pasunuru2017multi}
Ramakanth Pasunuru and Mohit Bansal. 2017.
\newblock Multi-task video captioning with video and entailment generation.
\newblock {\em arXiv preprint arXiv:1704.07489\/} .

\bibitem[{Pulijala et~al.(2013)Pulijala, Akula, and Syed}]{pulijala2013web}
Vasu Pulijala, Arjun~R Akula, and Azeemuddin Syed. 2013.
\newblock A web-based virtual laboratory for electromagnetic theory.
\newblock In {\em 2013 IEEE Fifth International Conference on Technology for
  Education (t4e 2013)\/}. IEEE, pages 13--18.

\bibitem[{R~Akula et~al.(2019)R~Akula, Todorovic, Y~Chai, and
  Zhu}]{r2019natural}
Arjun R~Akula, Sinisa Todorovic, Joyce Y~Chai, and Song-Chun Zhu. 2019.
\newblock Natural language interaction with explainable ai models.
\newblock In {\em Proceedings of the IEEE Conference on Computer Vision and
  Pattern Recognition Workshops\/}. pages 87--90.

\bibitem[{Ramanishka et~al.(2017)Ramanishka, Das, Zhang, and
  Saenko}]{ramanishka2017top}
Vasili Ramanishka, Abir Das, Jianming Zhang, and Kate Saenko. 2017.
\newblock Top-down visual saliency guided by captions.
\newblock In {\em Proceedings of the IEEE Conference on Computer Vision and
  Pattern Recognition (CVPR)\/}. volume~1, page~7.

\bibitem[{Reitter(2003)}]{reitter2003simple}
David Reitter. 2003.
\newblock Simple signals for complex rhetorics: On rhetorical analysis with
  rich-feature support vector models.
\newblock In {\em LDV Forum\/}. volume~18, pages 38--52.

\bibitem[{Simonyan and Zisserman(2014)}]{simonyan2014very}
Karen Simonyan and Andrew Zisserman. 2014.
\newblock Very deep convolutional networks for large-scale image recognition.
\newblock {\em arXiv preprint arXiv:1409.1556\/} .

\bibitem[{Soricut and Marcu(2003)}]{soricut2003sentence}
Radu Soricut and Daniel Marcu. 2003.
\newblock Sentence level discourse parsing using syntactic and lexical
  information.
\newblock In {\em Proceedings of the 2003 Conference of the North American
  Chapter of the Association for Computational Linguistics on Human Language
  Technology-Volume 1\/}. Association for Computational Linguistics, pages
  149--156.

\bibitem[{Sutskever et~al.(2014)Sutskever, Vinyals, and
  Le}]{sutskever2014sequence}
Ilya Sutskever, Oriol Vinyals, and Quoc~V Le. 2014.
\newblock Sequence to sequence learning with neural networks.
\newblock In {\em Advances in neural information processing systems\/}. pages
  3104--3112.

\bibitem[{Thomason et~al.(2014)Thomason, Venugopalan, Guadarrama, Saenko, and
  Mooney}]{thomason2014integrating}
Jesse Thomason, Subhashini Venugopalan, Sergio Guadarrama, Kate Saenko, and
  Raymond~J Mooney. 2014.
\newblock Integrating language and vision to generate natural language
  descriptions of videos in the wild.
\newblock In {\em Coling\/}. volume~2, page~9.

\bibitem[{Venugopalan et~al.(2016)Venugopalan, Hendricks, Mooney, and
  Saenko}]{venugopalan2016improving}
Subhashini Venugopalan, Lisa~Anne Hendricks, Raymond Mooney, and Kate Saenko.
  2016.
\newblock Improving lstm-based video description with linguistic knowledge
  mined from text.
\newblock {\em arXiv preprint arXiv:1604.01729\/} .

\bibitem[{Wang and Lan(2015)}]{wang2015refined}
Jianxiang Wang and Man Lan. 2015.
\newblock A refined end-to-end discourse parser.
\newblock In {\em Proceedings of the Nineteenth Conference on Computational
  Natural Language Learning-Shared Task\/}. pages 17--24.

\bibitem[{Yu et~al.(2016)Yu, Wang, Huang, Yang, and Xu}]{yu2016video}
Haonan Yu, Jiang Wang, Zhiheng Huang, Yi~Yang, and Wei Xu. 2016.
\newblock Video paragraph captioning using hierarchical recurrent neural
  networks.
\newblock In {\em Proceedings of the IEEE conference on computer vision and
  pattern recognition\/}. pages 4584--4593.

\end{thebibliography}

\end{document}